# The Role of Cognitive Modeling in Achieving Communicative Intentions

Marilyn A. Walker*
Mitsubishi Electric Research Laboratories
201 Broadway
Cambridge, MA 02139
USA
walker@merl.com

Owen Rambow[†]
Université Paris 7 and CoGenTex, Inc.
TALANA, UFR Linguistique, Case 7003
2, Place Jussieu
75251 Paris Cedex 05, France
rambow@linguist.jussieu.fr

## Abstract

A discourse planner for (task-oriented) dialogue must be able to make choices about whether relevant, but optional information (for example, the "satellites" in an RST-based planner) should be communicated. We claim that effective text planners must explicitly model aspects of the Hearer's cognitive state, such as what the hearer is attending to and what inferences the hearer can draw, in order to make these choices. We argue that a mere representation of the Hearer's knowledge is inadequate. We support this claim by (1) an analysis of naturally occurring dialogue, and (2) by simulating the generation of discourses in a situation in which we can vary the cognitive parameters of the hearer. Our results show that modeling cognitive state can lead to more effective discourses (measured with respect to a simple task).

## 1 Introduction

Text planning is the task for a speaker (S) of deciding what information to communicate to a hearer (H) and how and when to communicate it. Over the last few years a consensus has emerged that the text planning task should be formulated in terms of communicative goals or intentions [18, 24, 22]. Consider, for example, the RST-based planners developed at ISI [13, 20, 14]. These planners use the discourse relations proposed by Rhetorical Structure Theory (RST) [17] as plan operators, by interpreting the requirements on the related segments as preconditions, and the resultant effect of the discourse relation as a postcondition in a traditional AI planning architecture.

Two types (at least) of discourse relations have been identified in the literature. A SUBJECT-MATTER relation [17] or SEMANTIC relation [14] simply reflects a relation that exists independently in the world, such as causation. Each subject-matter relation can be seen as a rhetorical strategy for the linguistic realization of a range of communicative intentions [21]. A PRESENTATIONAL RELATION [17] or INTERPERSONAL relation [14] holds between two discourse segments such that the juxtaposition increases H's *strength* of belief, desire, or intention. Each presentational relation maps directly to a communicative intention. Examples of presentational relations include the MOTIVATION relation, which increases H's desire to perform an action, hopefully persuading H to form an intention to do the action.

Both subject-matter and presentational relations relate two clauses: (1) the NUCLEUS which realizes the main point; and (2) the SATELLITE which is auxiliary *optional* information. For example in the MOTIVATION relation shown in Figure 1, the SATELLITE is the belief which provides motivation to do the action realized by the proposal or suggestion in the NUCLEUS. Since the SATELLITE information may or may not be realized, previous text planners have run in either verbose or terse mode, in which either all or no satellite information is realized [21].

If an approach to text planning based on the notion of communicative intention is to succeed, it requires an appropriate representation of communicative goals, and of all mental states required for reasoning about these goals. This is especially true in the case of presentational relations. We can immediately observe that since such relations affect the degree of strength of H's belief, desire of intention, we need a gradual representation of mental attitudes. To our knowledge, no current text planner uses such a gradual representation. Second, it has been widely assumed that a model of what the hearer knows determines exactly when to include optional information in verbose mode: include optional information unless the hearer knows it. However, in our analysis of a corpus of 55 naturally-occurring dialogues, information that the hearer knew was frequently realized [34]. Consider the following short natural dialogue, part of a discussion about which Indian restaurant to go to for lunch:

(1) a. Listen to Ramesh.
    b. He's Indian.

Clearly, S wants to MOTIVATE H to accept his pro-

---

*Walker was partially funded by ARO grant DAAL03-89-C0031PRI and DARPA grant N00014-90-J-1863 at the University of Pennsylvania and by Hewlett Packard, U.K.

[†]Rambow was supported by NATO on a NATO/NSF postdoctoral fellowship in France.

posal with (1b). However, in this situation all of the discourse participants already knew that Ramesh was Indian. We hypothesize that Example (1) shows that there are cognitive processing motivations for S's choice to include information that is already known to the hearer, such as (1b), and that a model of H's cognitive processes are required for a text planner to appropriately decide when to include optional information.

The remainder of this paper is structured as follow. We start out by describing in more detail the problem facing text planners based on communicative goals (Section 2). In Section 3 we briefly review cognitive theories of deliberation and inference and relate these to an account of working memory. Next, in Section 4 we present the Design-World experimental environment, in which we embed our cognitive model. In Section 5, we present some examples of modeling experiments that suggest what sort of information S must access in order to generate efficient discourse. Finally, in Section 6 we briefly discuss possible implications for text planning architectures.

## 2 Achieving Communicative Goals

In this paper, we focus on presentational relations, and we use the MOTIVATION relation as a prototypical presentational discourse relation to illustrate our points. The MOTIVATION relation, whose effect is to increase H's desire to perform an action, is shown in Figure 1. We will call the nucleus of MOTIVATION the PROPOSAL, and any information that can serve as the satellite of MOTIVATION we will call a WARRANT.

What sort of representation is needed in order to use MOTIVATION for discourse planning? The effect of presentational relations is always to *increase* H's belief, desire, or intention. Thus we will need (in the case of MOTIVATION) some sort of representation of *degree* of desire. In a first attempt at using MOTIVATION, we will use utility theory [6] and simply associate utilities with proposed actions. Under this view, an agent's strength of desire to perform an action is the utility he or she believes performing the action will yield, where "utility" is a quantifiable variable. In Section 6 we will discuss the limitations of this approach.

However, even though utility theory can be used as the theoretical underpinning of the MOTIVATION relation, it will not in general be sufficient because it does not take into account the way in which H's beliefs interact with his attentional state, and the way that H's cognitive limitations interact with the demands of the task.

Consider the following scenario. An agent S wants an agent H to accept a proposal P. In the situation where H always deliberates and H knows of options which compete with proposal P, H cannot decide whether to accept P without a warrant.[1] Previous work has assumed that the warrant can be omitted if it is already believed by H. Presumably the speaker in (2) will not say *It's shorter* if she believes that the hearer knows that the Walnut St. route was shorter.

(2) a. Let's walk along Walnut St.

b. It's shorter.

However, consider again (1), repeated here as (3):

(3) a. Listen to Ramesh.

b. He's Indian.

The warrant in (3b) was included despite the fact that it was common knowledge among the conversants.[2] Our hypothesis is that 3 shows that speakers distinguish between information that H *knows* and information that is *salient* for H [27]. Thus even if H knows a warrant for adopting S's proposal, if the warrant is not salient for H, then S may choose to include it with a proposal.

We define SALIENT as available in current Working Memory, referred to below as Attention/Working Memory or AWM. A model of H's attentional state will distinguish between those discourse entities and beliefs which are currently available in working memory, and thus salient, and those that are not. In Section 3, we introduce an operationalization of AWM and discuss how S can model what is salient for H.

When a warrant is not SALIENT, H must either infer the warrant information, or retrieve it from long term memory, or obtain it from an external source in order to evaluate S's proposal. Thus S's communicative choice as to whether to include the warrant satellite may depend on S's model of H's attentional state. Furthermore, it is possible that, even though the warrant is not salient, merely a trivial processing effort is required to retrieve it, so that it is not worth saying. Another possibility is that processing the warrant utterance requires effort that can be traded off against other processes such as retrieval and inference. In other words, S may decide that it is easier just to say the warrant rather than require H to infer or retrieve it. We will call a text planning strategy that always includes the warrant satellite the Explicit-Warrant strategy.

We see that in addition to S modeling H's knowledge, H's attentional state and expectations about other aspects of H's cognitive processes may also influence S's text planning decisions, and S cannot simply represent H's beliefs as a set of pairs of propositions and associated utility.[3] In sum, the choice is hypothesized to depend on cognitive properties of H, e.g. what H knows, H's attentional state, and H's processing capabilities, as well as properties of the task and the communication channel.

---

[1] Elsewhere we consider scenarios in which H always accepts S's proposal without a warrant and in which H never knows of competing options to P [35].

[2] After inferring the intended relation the hearer still must decide whether s/he believes that Indians know of good Indian restaurants [37].

[3] How S would have access to the necessary information is a separate issue, briefly discussed in Section 6.

| | |
|---|---|
| relation name: | MOTIVATION |
| constraints on Nucleus: | presents an action in which H is the actor unrealized with respect to the context of N (a proposal in our terminology) |
| constraints on Satellite: | none |
| constraints on the Nucleus + Satellite combination: | H's comprehending the Satellite increases H's desire to perform action presented in the Nucleus |
| the effect: | H's desire to perform action presented in the Nucleus is increased |

Figure 1: RST definition of Motivation Relation

In this paper, we explore some cognitive issues involved in planning to include satellite information found in RST presentational relations and the representational demands that arise for text planning tasks. We will argue that S must model H's cognitive state in a much more detailed manner than previously assumed and put forth a proposal about how S might access the information required in order to do so. In order to provide evidence for our claim, we will use the cognitive modeling methodology developed in [34], in which communicative strategies on the part of S can be represented and their effects can be empirically tested. This architecture allows us to identify parameters in H's cognitive state that affect S's communicative decisions and therefore must be modeled. The simulation/modeling environment is called Design-World.

## 3 Modeling Cognitive Limits

In Section 2 we proposed some cognitive factors, motivated by proposals in naturally occurring dialogue, that may provide limits on whether agents can optimally deliberate proposed actions or make inferences. We hypothesized that these factors will determine when Explicit-Warrant is an effective strategy. Here we briefly present a way of cognitively modeling agents' limited attention and the relationship of limited attention to deliberation and inference.

It is well known that human agents have limited attentional capacity [19] and it has been argued that limited attention plays a major role in theoretical and scientific reasoning [28, 15, 31], ie. in deduction, and belief and intention deliberation. We hypothesized that Example (2) shows that a warrant must be SALIENT for both agents in order to be used in deliberation, i.e. for it to motivate H effectively. This fits with the psychological theories mentioned above, that only salient beliefs are used in deliberation and inference.

In Design-World, salience is modeled by the AWM model, adapted from [16]. While the AWM model is extremely simple, Landauer shows that it can be parameterized to fit many empirical results on human memory and learning [2]. AWM consists of a three dimensional space in which propositions acquired from perceiving the world are stored in chronological sequence according to the location of a moving memory pointer. The sequence of memory loci used for storage constitutes a random walk through memory with each locus a fixed distance from the previous one. If items are encountered multiple times, they are stored multiple times [12].

When an agent retrieves items from memory, search starts from the current pointer location and spreads out in a spherical fashion. Search is restricted to a particular search radius: radius is defined in Hamming distance. For example if the current memory pointer locus is (0 0 0), the loci distance 1 away would be (0 1 0), (0 -1 0), (0 0 1), (0 0 -1), (-1 0 0), and (1 0 0). The actual locations are calculated modulo the memory size. The limit on the search radius defines the capacity of attention/working memory and hence defines which stored beliefs and intentions are SALIENT.

The radius of the search sphere in the AWM model is used as the parameter for Design-World agents' resource-bound on attentional capacity. In the experiments below, memory is 16x16x16 and the radius parameter varies between 1 and 16. Agents with an AWM of 1 have access to 7 loci, and since propositions are stored sparsely, they only remember the last few propositions that they acquired from perception. Agents with an AWM of 16 can access everything they know.[4]

The AWM model also gives us a way to measure (1) the number of retrievals from memory in terms of the number of locations searched to find a proposition; (2) the number of inferences that the agents make as they means-end reason and draw content-based inferences; and (3) the number of messages that the agents send to one another as they carry out the dialogue. The amount of effort required for each of these cognitive processes are parameters of the model. These cost parameters support modeling various cognitive or text planning architectures, e.g. varying the cost of retrieval models different assumptions about memory. Since these cognitive processes are the primitives involved in text planning, this framework can be used to model many different architectures rather than the results being specific to a particular text-planning architecture.

---
[4]The size of memory was determined as adequate for producing the desired level of variation in the current task across all the experimental variables, while still making it possible to run a large number of simulations involving agents with access to all of their memory in a reasonable amount of time. In order to use the AWM model in a different task, the experimenter might want to explore different sizes for memory.

The retrieval parameter alone allows us to model many different assumptions about memory. For example, if retrieval is free then all items in working memory are instantly accessible, as they would be if they were stored in registers with fast parallel access. If AWM is set to 16, but retrieval isn't free, the model approximates slow spreading activation that is quite effortful, yet the agent still has the *ability* to access all of memory, given enough time. If AWM is set lower than 16 and retrieval isn't free, then we model slow spreading activation with a timeout when effort exceeds a certain amount, so that an agent does not have the *ability* to access all of memory. Thus the AWM parameter supports a distinction between an agent's ability to access all the information stored in its memory, and the effort involved in doing so.

It does not make sense to fix absolute values for the retrieval, inference and communication cost parameters in relation to human processing. However, Design-World supports exploring issues about the *relative* costs of various processes. These relative costs might vary depending on the language that the agents are communicating with, properties of the communication channel, how smart the agents are, how much time they have, and what the demands of the task are [23]. Below we vary the relative cost of communication and retrieval.

The advantages of the AWM model is that it has been shown to reproduce, in simulation, many results on human memory and learning. Because search starts from the current pointer location, items that have been stored most recently are more likely to be retrieved, predicting recency effects [2]. Because items that are stored in multiple locations are more likely to be retrieved, the model predicts frequency effects [16]. Because items are stored in chronological sequence, the model produces natural associativity effects [1].

The overall agent architecture is modeled on the IRMA agent architecture [3] with the addition of AWM. See Figure 2. As Figure 2 shows, AWM interacts with other processing because deliberation and means-end reasoning only operate on salient beliefs. This means that limits in AWM produces a concomitant inferential limitation, i.e. if a belief is not salient it cannot be used in deliberation or means-end-reasoning. Thus mistakes that agents make in their planning process have a plausible cognitive basis. Agents can both fail to access a belief that would allow them to produce an optimal plan, as well as make a mistake in planning if a belief about how the world has changed as a result of planning is not salient. Depending on the preceding discourse, and the agent's attentional capacity, the propositions that an agent *knows* may or may not be *salient* when a proposal is made [27].

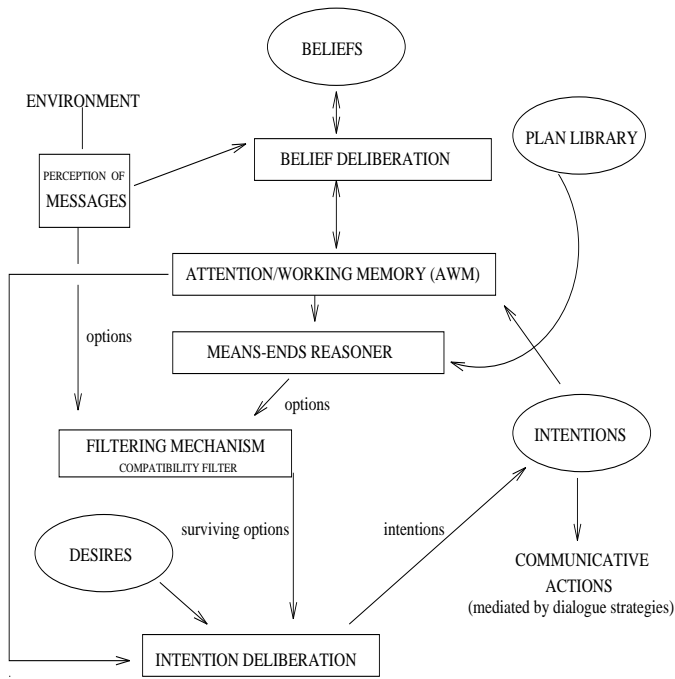

Figure 2: Design-World version of the IRMA Agent Architecture for Resource-Bounded Agents with Limited Attention (AWM)

## 4 Experimental Environment: Design-World

Design-World is an experimental environment for testing the relationship between ways of realizing communicative intentions and agents' cognitive capabilities, similar to the single-agent TileWorld simulation environment [25, 11]. Design-World agents can be parametrized as to discourse strategy, e.g. whether to use the Explicit-Warrant strategy, and the effects of this strategy can be measured against a range of cognitive and task parameters. In Section 4.1, we describe the Design-World domain and task. In Section 4.2, we describe two alternate discourse strategies. In Section 4.3, we discuss how performance is evaluated and compared. Finally, in Section 5 we present the experimental results.

### 4.1 Design World Domain and Task

The Design-World task requires two agents to carry out a dialogue in order to negotiate an agreement on the design of the floor plan of a two room house [29, 38]. The DESIGN-HOUSE plan requires the agents to agree on how to DESIGN-ROOM-1 and DESIGN-ROOM-2. At the beginning of the simulation, both agents know the structure of the DESIGN-HOUSE plan. Each agent has 12 items of furniture that can be used in the plan, with utility scores ranging from 10 to 56. A potential final design plan negotiated via a (simulated) dialogue is shown in Figure 3.

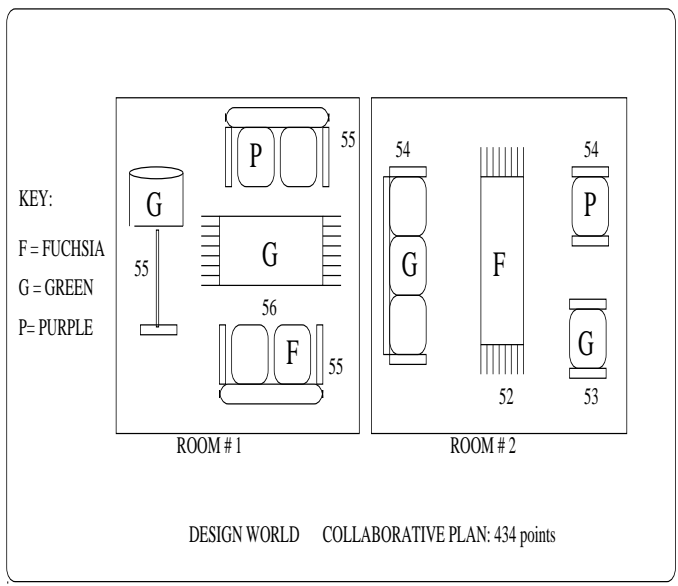

Figure 3: One Final State for Design-World Standard Task: Represents the Collaborative Plan Achieved by the Dialogue, 434 points

Negotiating an agreement consists of a cycle in which: (1) individual agents perform means-end reasoning about OPTIONS to use various furniture pieces that they have in the floor plan; (2) individual agents deliberate about which options are preferable; (3) then agents use the OPTIONS identified as the content of PROPOSALS to other agents to PUT a piece of furniture into one of the rooms; (4) then these proposals are ACCEPTED or REJECTED by the other agent, or acceptance/rejection is postponed by a CLARIFICATION.

Dialogue (4) illustrates agents' communication for part of designing room-1. Both the artificial language that the agents communicate with and a gloss generated from that language in *italics* are shown:[5]

(4)  1: BILL: *Let's put the green rug in the study.*
    (propose agent-bill agent-kim option-10: put-act
    (agent-bill green rug room-1))

  2: KIM: *Then, let's put the green lamp there.*
    (propose agent-kim agent-bill option-33: put-act
    (agent-kim green lamp room-1))

  3: BILL: *Next, let's put the green couch in the study.*
    (propose agent-bill agent-kim option-45: put-act
    (agent-bill green couch room-1))

  4: KIM: *No, instead let's put the purple couch in the study.*
    (reject agent-kim agent-bill option-56: put-act (agent-kim purple couch room-1))

On receiving a proposal, an agent deliberates whether to ACCEPT or REJECT the proposal [6]. Each furniture item has a value that contributes to an evaluation of the final plan. The values on the furniture items range from 10 to 56, and both agents' furniture items range over these values. Agents know the values of all the furniture items at the beginning of the dialogue. The values of the furniture items are used to MOTIVATE the agents to ACCEPT a proposal, as well as providing a way of objectively evaluating agents' performance. In other words, we define each potential action to have an associated SCORE; agents can evaluate whether their desire to do an action is increased by comparing the score of the proposed action with other actions that they know about.

For example, at the beginning of the dialogue, Agent-Kim has stored in memory the proposition that (score green-rug 56). When she receives Bill's proposal as shown in (4-1), she evaluates that proposal in order to decide whether to accept or reject it. As part of evaluating the proposal she will attempt to retrieve the score proposition stored earlier in memory. Thus the propositions about the scores of furniture items are WARRANTS for supporting deliberation.

Agents REJECT a proposal if deliberation leads them to believe that they know of a better option. For example, in (4-4) Kim rejects the proposal in (4-3), for pursuing option-45, and proposes option-56 instead. The form of the rejection as a counter-proposal is based on observations about how rejection is communicated in naturally-occurring dialogue as codified in the COLLABORATIVE PLANNING PRINCIPLES [36].

Proposals 1 and 2 are inferred to be implicitly ACCEPTED because they are not rejected [36]. If a proposal is ACCEPTED, either implicitly or explicitly, then the option that was the content of the proposal becomes a mutual intention that contributes to the final design plan [26, 33, 29].

The model of AWM discussed above plays a critical role in determining agents' performance. Remember that only *salient* beliefs can be used in means-end reasoning and deliberation, so that if the warrant for a proposal is not salient, the agent cannot properly evaluate a proposal.

### 4.2 Varying Discourse Strategies

Agents are parametrized for different discourse strategies by placing different expansions of discourse plans in their plan libraries. In Design-World the only discourse plans required are plans for PROPOSAL, REJECTION, ACCEPTANCE, CLARIFICATION, OPENING and CLOSING. The only variations discussed here are variations in the expansions of PROPOSALS.

The All-Implicit strategy is an expansion of a discourse plan to make a PROPOSAL, in which a PROPOSAL decomposes trivially to the communicative act of PROPOSE. In dialogue (4), both Design-World agents communicate using the All-Implicit strategy, and proposals are expanded to the PROPOSE communicative acts shown in 1, 2, and 3 in dialogue (4). The All-Implicit strategy never includes warrants in proposals, leaving it up to the other agent to retrieve them from memory.

---
[5]The generation of the gloss was not a focus of this study and was done via adhoc methods.

The Explicit-Warrant strategy expands the PROPOSAL discourse act to be a WARRANT followed by a PROPOSE utterance [32].[6] Since agents already know the point values for pieces of furniture, warrants are always optional in the experiments here. In RST terms, an agent with the Explicit-Warrant strategy always chooses to MOTIVATE every proposal. For example (5-1) is a WARRANT for the proposal in (5-2):

(5) 1: TED: **Putting in the green rug is worth 56.**
(say agent-ted agent-ben bel-10: score (option-10: put-act (agent-ted green rug room-1) 56))

2: TED: *So, let's put the green rug in the study.*
(propose agent-ted agent-ben option-10: put-act (agent-ted green rug room-1))

3: BEN: **Putting in the green lamp is worth 55.**
(say agent-ben agent-ted bel-34: score (option-33: put-act (agent-ben green lamp room-1) 55))

4: BEN: *So, let's put the green lamp in the study.*
(propose agent-ben agent-ted option-33: put-act (agent-ben green lamp room-1))

The fact that the green rug is worth 56 points is motivation for adopting the intention of putting the green rug in the study. Whether it is good motivation depends on what other options the agent knows about and what their utilities are. The Explicit-Warrant strategy models naturally occurring examples such as those in 2 because the score information used by the hearer to deliberate whether to accept or reject the proposal is already mutually believed.

### 4.3 Evaluating Performance

Remember that we incorporate cognitive modeling into Design-World so that attentional capacity and the cost of various cognitive processes are parameters. To evaluate PERFORMANCE, we compare the Explicit-Warrant strategy with the All-Implicit strategy while we vary agents' attentional capacity, and the cost of retrieval, inference and communication. Evaluation of the resulting DESIGN-HOUSE plan is parametrized by (1) COMMCOST: cost of sending a message; (2) INFCOST: cost of inference; and (3) RETCOST: cost of retrieval from memory:

PERFORMANCE = Task Defined RAW SCORE
− (COMMCOST × total messages)
− (INFCOST × total inferences)
− (RETCOST × total retrievals)

RAW SCORE is task specific: in the Standard task we simply summarize the point values of the furniture pieces in each PUT-ACT in the final design.

---

[6]The ordering of these two acts as given lets us have a simple control regime for processing utterances. The reverse ordering would require the agent to check whether it has more messages before processing the current message.

We simulate 100 dialogues at each parameter setting for each strategy. Because the AWM model is probabilistic, the agents do not perform identically on each trial, and their performance over the 100 dialogues defines a performance distribution. In order to compare two strategies, we test whether the differences in the performance distributions are significant, using the Kolmogorov-Smirnov (KS) two sample test [30].

A strategy A is BENEFICIAL as compared to a strategy B, for a set of fixed parameter settings, if the difference in distributions using the Kolmogorov-Smirnov two sample test is significant at $p < .05$, in the positive direction, for two or more AWM settings.

A strategy is DETRIMENTAL if the differences go in the negative direction. Strategies may be neither BENEFICIAL or DETRIMENTAL, there may be no difference between two strategies.

## 5 Experimental Results on Providing Motivation

This section discusses a few experimental results on the Explicit-Warrant discourse strategy, which we compare with the All-Implicit strategy. Here we simply test the effect of whether the warrant is salient or not, whether there is any processing effort associated with retrieving the warrant from long term memory, and whether the cost of processing an utterance is high.

Differences in performance between the Explicit-Warrant strategy and the All-Implicit strategy are shown via a DIFFERENCE PLOT such as Figure 4. In Figure 4 performance differences are plotted on the Y-axis and AWM settings are shown on the X-axis. If the plot is above the dotted line for 2 or more AWM settings, then the Explicit-Warrant strategy may be BENEFICIAL. Each point represents the difference in the means of 100 runs of each strategy at a particular AWM setting.

### 5.1 Explicit Warrant reduces Retrievals

Dialogues in which one or both agents use the Explicit-Warrant strategy are more efficient when retrieval has a cost.

Figure 4 shows that the Explicit-Warrant strategy is detrimental at AWM of 3,4,5 for the Standard task, in comparison with the All-Implicit strategy, if retrieval from memory is free (KS 3,4,5 > .19, $p < .05$). This is because making the warrant salient displaces information about other pieces when agents are attention-limited.

However, Figure 5 shows that Explicit-Warrant is beneficial when retrieval has an associated processing cost. By AWM values of 3, performance with Explicit-Warrant is better than All-Implicit because the warrants intended to motivate the hearer and used by the hearer in deliberation are made salient with each proposal (KS for AWM of 3 and above > .23, $p < .01$). At AWM parameter settings of 16, where agents have the ability to search a huge belief space for beliefs to be used as warrants, the saving in processing time is substantial. Again at the lowest AWM settings, the strategy is not beneficial because it displaces information about other pieces from AWM. However in Figure 5, in contrast with Figure 4, retrieval has an associated cost. Thus the savings in retrieval balance out with the loss of raw score so that the strategy is not DETRIMENTAL.

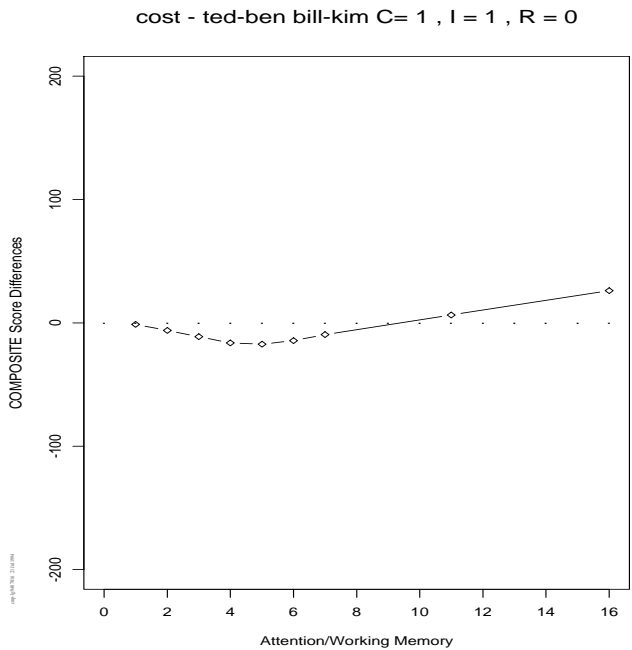
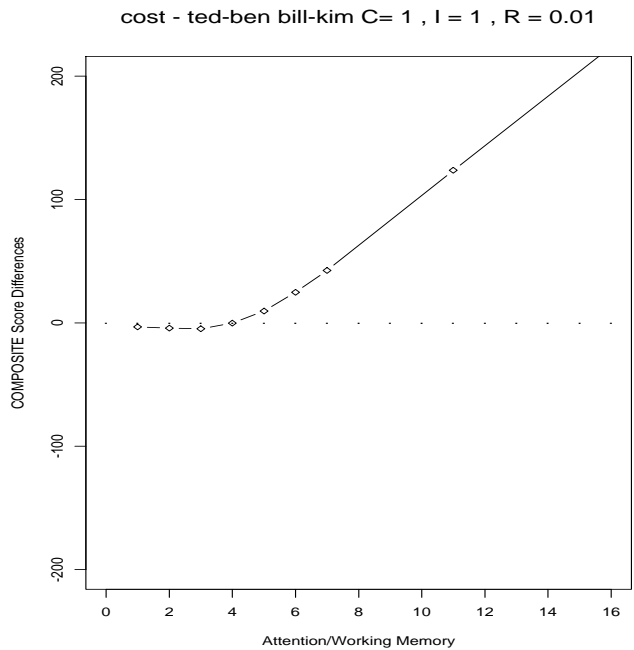

Figure 4: If Retrieval is Free, Explicit-Warrant is detrimental at AWM of 3,4,5: Strategy 1 of two Explicit-Warrant agents and strategy 2 of two All-Implicit agents: Task = Standard, commcost = 1, infcost = 1, retcost = 0

Figure 5: Retrieval costs: Strategy 1 is two Explicit-Warrant agents and strategy 2 is two All-Implicit agents: Task = Standard, commcost = 1, infcost = 1, retcost = .01

### 5.2 Explicit Warrant is detrimental if Communication is Expensive

Finally we can amplify the results shown in Figure 4 by positing that in addition to there being no processing effort for retrieving from memory, processing the additional warrant utterance requires a lot of processing effort. Figure 6 shows that if communication cost is 10, and inference and retrieval are free, then the Explicit-Warrant strategy is DETRIMENTAL (KS for AWM 1 to 5 > .23, p < .01). This is because the Explicit-Warrant strategy increases the number of utterances required to perform the task; it doubles the number of messages in every proposal. If communication is expensive compared to retrieval, processing additional warrant utterances is highly detrimental if there would be no effort involved in retrieving them, i.e., if they are essentially already salient.

## 6 Implications for Text Planning

The experiments reported in the previous section show that there is a direct relation between H's attentional state and the advisability of including warrants in a text plan. There are two ways in which we have modeled different aspects of H's attentional state in the experiments reported in the previous sections: we have varied the cost of retrieval, and we have varied the size of AWM. We can think of H's attentional state as comprising a (small) active working memory, and a larger long-term memory. We vary whether an agent has the *ability* to retrieve an item by varying the radius of AWM. We vary the amount of *effort* involved in retrieving an item by varying the cost of retrieval [16, 23, 2]. If a warrant for the proposal is in short-term memory and can be accessed virtually cost-free, then Figure 4 shows that generating the warrant explicitly can actually be detrimental, since it can displace other information. The effect is further magnified if communication is very costly, as shown in Figure 6. (Cost of communication may by increased by a large number of factors, such as the linguistic complexity of the generated message, the fact that H is not fully competent in the language of communication, or noise in the channel of communication.) Thus, if S knows that information that can serve as a warrant is salient to H, then no warrant should be generated. On the other hand, if a cost is in fact associated with retrieval, as in the experiment reported

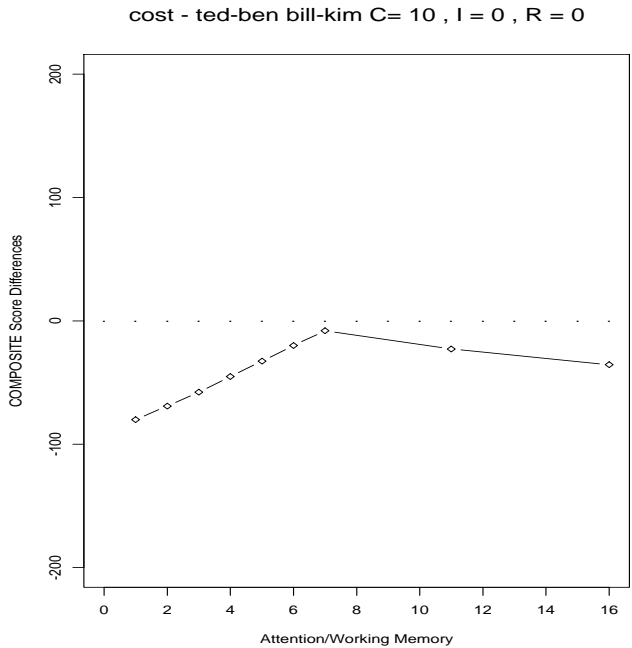

Figure 6: If Communication is Expensive: Communication costs can dominate other costs in dialogues. Strategy 1 is two Explicit-Warrant agents and strategy 2 is two All-Implicit agents: Task = Standard, commcost = 10, infcost = 0, retcost = 0

in Figure 5, then we can see that generating a warrant is beneficial, especially as the size of AWM increases and agents have the ability to access all of long term memory.

We conclude that for a text planner to decide whether or not to communicate certain non-essential information, such as a warrant as part of a MOTIVATE relation, depends not just on the effect of the relation (which must of course match the discourse goal), but also on the attentional state of H, and on the other factors such as the cost of communication. A maximally efficient text planner will need to have access to:

- a model of H's attentional state;
- an algorithm that, given the attentional state model and additional parameters such as the costs of communication and retrieval determines whether H knows accessible information which can serve as a warrant.

Here "accessible information" means either that the information is already salient or that it can be retrieved at a reasonable cost, given the costs of communication and retrieval. If we assume that the algorithm defines a binary predicate **NOT-ACCESSIBLE**, we can formalize the RST relation MOTIVATE as a plan operator **Motivation** as given in Figure 7. The format follows the format of the plan operators given in [21], except that, for simplicity, we conflate the intentional and the rhetorical levels.[7] This plan operator is of course only meant to be suggestive, and we are not committed to any details.

Of course, procedure **NOT-ACCESSIBLE** crucially relies on a proper model of H's attentional state and on an algorithm that accesses it. We intend to investigate these issues in future work, but sketch some possible solutions here. An obvious candidate for the model is the AWM model used in the simulations itself. (In fact, it is quite plausible that speakers use their own attentional state as a model for that of the hearer.) The algorithm could then be defined very straightforwardly in terms of a three-dimensional boolean matrix, indexed on distance in memory, communication cost, and retrieval cost. The value for a given triple indicates whether or not information stored at this distance is accessible. The values in the table are determined using the simulation environment. Presumably, this process weakly corresponds to the acquisition of proper text planning strategies by human agents.

While the AWM model is an obvious candidate for the model of H's attentional state, certain aspects of this model do not exactly match some widely believed and intuitively motivated observations about hierarchical discourse structure [10, 17]. However, hierarchical structure interacts with attentional state in ways that have not been fully explored in the literature to date. In particular, if a discourse segment consists of a nucleus with a hierarchically complex satellite that is extremely long, then a further satellite to the same nucleus may well require repetition of the nucleus [34]. Neither RST nor the model of Grosz and Sidner accounts for such effects. We conclude that it is not a priori obvious that hierarchical structure contradicts our model. We will investigate this issue further.

Throughout this paper, we have used the MOTIVATE relation in order to motivate our claims. However, similar observations apply to other presentational relations, such as BACKGROUND, or EVIDENCE as shown in (6):[8]

(6) a. Clinton has to take a stand on abortion rights for poor women.

   b. He's the president.

Here (6b) is already known to the discourse participants, but saying it makes it salient. A discussion of the EVIDENCE relation, however, is complicated by the need to find a proper representation of the degree of strength of belief, since utility theory is not appropriate as a representation of degree of belief [9]. In other work, we have developed a version of Gallier's theory of belief revision which takes into account the effect of limited working memory on agent's belief revision processes [7, 8, 5, 34]. This theory could be used for the RST EVIDENCE relation.

---

[7] Moore and Paris argue that for presentational relations (such as MOTIVATE), there is a one-to-one mapping between intentional and rhetorical structure. Therefore, conflating them is theoretically justified here.

[8] We also believe that whether a known proposition is salient is an issue for supporting content-based inferences, and thus cognitive modeling may be required for text planning of subject-matter relations as well.

| NAME: | MOTIVATION |
|---|---|
| EFFECT: | (DESIRE ?hearer (DO ?hearer ?act) ?utility-act) |
| CONSTRAINTS: | (AND (AGENT ?act ?hearer) |
| | (UNREALIZED ?act) |
| | (NOT-ACCESSIBLE ?hearer (UTILITY ?act ?utility-act))) |
| SATELLITE | (BEL ?hearer (UTILITY ?act ?utility-act)) |
| NUCLEUS: | (BEL ?hearer (WANT ?speaker (DO ?hearer ?act))) |

Figure 7: The MOTIVATION plan operator

However, the use of a simple evaluation function for the representation of gradual strengths (of desire, belief, etc.) is in itself problematic. In this paper, we used the theoretical construct of utility as the basis for degree of desire for the MOTIVATION relation. However, observations of human dialogue show that there are many evaluation functions in the real world that can be the basis for the MOTIVATION relation. Furthermore, these evaluation functions are incomparable and competing, as shown by (7), asserted by a speaker while walking to work:

(7) I don't like going down that way.
It may be shorter, but I don't like it.

The speaker's desire for an aesthetic environment on her walk has in this case overridden her desire for the fastest route to work. However if she were late, the efficiency evaluation function might dominate. In the real world or in real text planning domains, the issue of multiple competing evaluation functions on potential intended actions must be addressed. Observe that this issue is crucially related to the proper theoretical discussion of presentational relations.

Finally, we would like to observe that there is evidence that the results presented here are domain-independent. The task that the agents are performing in Design-World is a simple task without any complex constraints, which we would expect to be a subcomponent of many other tasks. The model of limited resources we used was cognitively based, but the cost parameters allow us to model different agent architectures, and we explored the effects of different cost parameters. The Explicit-Warrant strategy is based on simple relationships between different facts which we would expect to occur in any domain, i.e. the fact that some belief can be used as a WARRANT for accepting a proposal should occur in almost any task. Furthermore, our results are confirmed by naturally occurring discourses in a wide variety of domains [37, 4, 34].

## 7 Conclusion

In this paper, we have argued that a text planner that is based on the notion of communicative intention and on plan operators that explictly represent such intentions must also incorporate a sophisticated model of the hearer's attentional state, and the ability to use this model in order to make decisions about whether or not to include optional information (satellites of presentational relations). We have motivated this claim using naturally occurring dialogues, and by experimental results from a simulation environment which implements a simple, but psychologically plausible model of attentional state. Future work includes extending the analysis to the whole range of presentational relations, defining precisely a hearer model that can be used in text planning and an associated algorithm that can be used by the plan operators, and a theoretical investigation of the interaction between textual hierarchy and attentional state.